\documentclass[conference]{IEEEtran}
\IEEEoverridecommandlockouts

\usepackage{cite}
\usepackage{amsmath,amssymb,amsfonts}
\usepackage{algorithm}
\usepackage{algpseudocode}
\makeatletter
\def\ALG@name{Algorithm}
\makeatother

\usepackage{graphicx}
\usepackage{textcomp}
\usepackage{booktabs}
\usepackage{multirow}
\usepackage{siunitx}
\usepackage{stfloats}
\usepackage{balance}
\usepackage{xspace}
\usepackage{url}
\usepackage{algpseudocode}
\usepackage{array}
\usepackage{makecell}
\usepackage{tabularx}
\newcolumntype{Y}{>{\centering\arraybackslash}X}
\usepackage[T1]{fontenc}
\usepackage{microtype}
\usepackage{algorithm}
\usepackage{algpseudocode}

\usepackage{amsmath}
\usepackage{amssymb}
\usepackage{bm}

\usepackage{xcolor}

\definecolor{appleorange}{RGB}{255,149,0}
\usepackage[table]{xcolor}
\definecolor{bestmetric}{RGB}{255, 239, 213}   
\definecolor{best3d}{RGB}{219, 235, 255}       
\definecolor{bestcommercial}{RGB}{226, 255, 219}  
\definecolor{bestoverall}{RGB}{255, 245, 200} 

\definecolor{applegray}{RGB}{142,142,147}
\definecolor{applelightgray}{RGB}{199,199,204}

\usepackage{hyperref}

\newcommand{\iris}{IRIS\xspace}

\newcommand{\colorsquare}[1]{%
  \raisebox{0.3em}{\colorbox{#1}{\hspace{0.3em}}}%
}

\begin{document}

\title{\iris: Learning-Driven Task-Specific Cinema Robot Arm for Visuomotor Motion Control
\thanks{$^{1}$This work was conducted while Qilong Cheng was at the University of Toronto.}
}

\author{
\IEEEauthorblockN{Qilong Cheng}
\IEEEauthorblockA{Mechanical and Aerospace Eng. Department\\
New York University\\
\texttt{qc1007@nyu.edu}}
\and
\IEEEauthorblockN{Matthew Mackay}
\IEEEauthorblockA{MIE Department\\
University of Toronto\\
\texttt{mackay@mie.utoronto.ca}
}
\and
\IEEEauthorblockN{Ali Bereyhi}
\IEEEauthorblockA{ECE Department\\
University of Toronto\\
\texttt{ali.bereyhi@utoronto.ca}}
}

\maketitle

\begin{abstract}
Robotic camera systems enable dynamic and repeatable motion beyond human capabilities.  
%
Yet their adoption is restricted by high costs and operational complexity of industrial-grade hardware.
We present \textit{intelligent robotic imaging system} (\iris), a task-specific 6-DOF manipulator, designed for autonomous learning-driven cinematic motion control.
%
Our system leverages a vertically integrated stack that combines a lightweight, 3D-printed hardware design with a visuomotor imitation learning framework. 
By employing a goal-conditioned adaptation of Action Chunking with Transformers (ACT), \iris learns to execute object-aware, perceptually smooth trajectories directly from human expert demonstrations, without the need for explicit geometric programming.
The complete system costs under \$1,000\,USD, supports a 1.5\,kg payload, and achieves approximately 1\,mm repeatability. 
%
Real-world experiments demonstrate accurate tracking and reliable autonomous execution that generalizes across diverse cinematic trajectories.
Implementation details for \iris{} are available at: \url{https://github.com/thejerrycheng/iris}.
%
%
\end{abstract}

\begin{IEEEkeywords}
robotic motion control, imitation learning, visuomotor control, full-stack robotic systems
\end{IEEEkeywords}


\section{Introduction}
A \emph{cinema robot} is a robotic motion-control system, typically a multi-degree-of-freedom (multi-DOF) arm or rig, designed to execute smooth, precise, and repeatable camera motion for shot execution and visual-effects continuity \cite{atherton2019robotic, fielding1985technique}.
Such systems are widely used in film, commercials and television production to capture dynamic, high-speed, or multi-pass shots that demand sub-millimeter repeatability and precise temporal alignment \cite{atherton2019robotic, delacruz2016robots}.
State-of-the-art cinema robots achieve high speed, large payload capacity, and high repeatability, but typically rely on industrial-grade actuation and control architectures that incur substantial cost and operational complexity \cite{atherton2019robotic,mrmc_faq_motion_control, mpstudio_kira}.
As a result, these platforms remain largely inaccessible to independent creators and research environments.
In contrast, recent low-cost and open-source robotic manipulators demonstrate the feasibility of 3D-printed hardware for learning-based control, but often trade off workspace, speed, or structural rigidity, limiting their suitability for professional cinematic camera motion \cite{mick2019reachy, openarm, balasubramanian2015analysis}.


\begin{figure}[t]
    \centering
    \includegraphics[width=\linewidth]{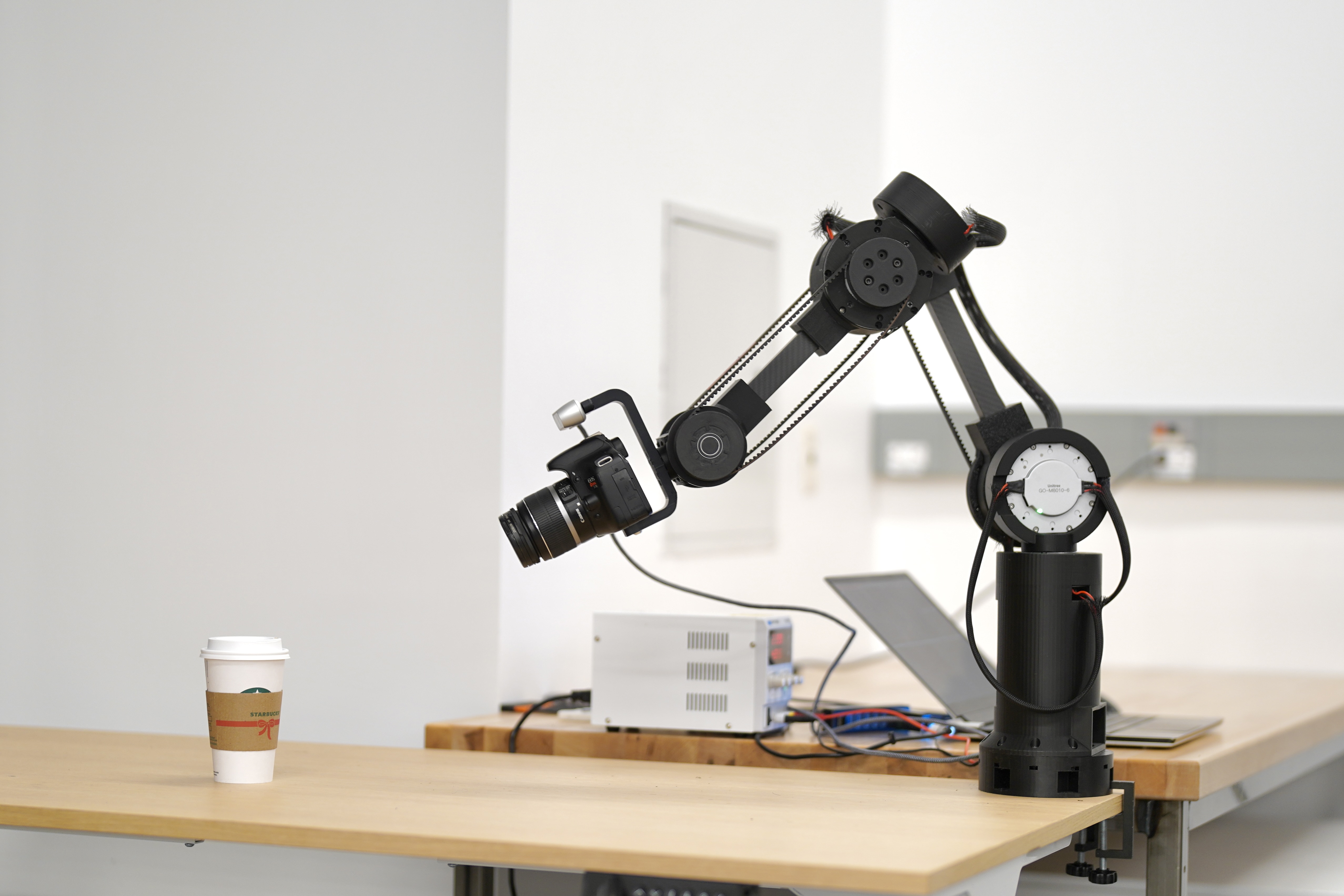}
    \caption{Tabletop deployment of the \iris{} prototype performing a real-world demonstration. An end-effector-mounted camera captures a target object (cup) using visuomotor control.}
    \label{fig:intro}
    \vspace{-1em}
\end{figure}

%
This study explores a task-specific alternative: \emph{what if a robotic arm were designed exclusively for cinematic motion, and learned camera behavior directly from expert operators?}
Our goal is to lower the barrier to professional camera motion control, and enable broader access for creators and researchers alike. 
We present \textit{Intelligent Robotic Imaging System}, \textbf{\iris}, a compact, low-cost, 3D-printed 6-DOF cinema robot designed explicitly for dynamic camera motion rather than adapted from industrial manipulators. 
\autoref{fig:intro} shows the hardware prototype of \iris in a tabletop setup. 
The key contributions of this work are as follows.
\begin{itemize}
    \item \textbf{Task-specific hardware design.} A low-cost 3D-printed 6-DOF manipulator is co-designed to meet the reachability, accuracy, and repeatability for cinema motion control.
    \item \textbf{Goal-conditioned ACT learning.} An imitation learning framework is developed, which extends Action Chunking with Transformers (ACT) with visual goal conditioning to generate obstacle-aware, goal-directed camera motion directly from observations.
    \item \textbf{End-to-end real-world validation.} A vertically integrated system is implemented and validated on physical hardware, showing accurate low-level control and autonomous high-level motion control.
\end{itemize}


To the best of our knowledge, \iris{} is among the first systems to prioritize low-cost cinematic shot automation through joint hardware–control co-design driven by robot learning. 

\begin{figure*}[t!]
    \centering
    \includegraphics[width=\linewidth]{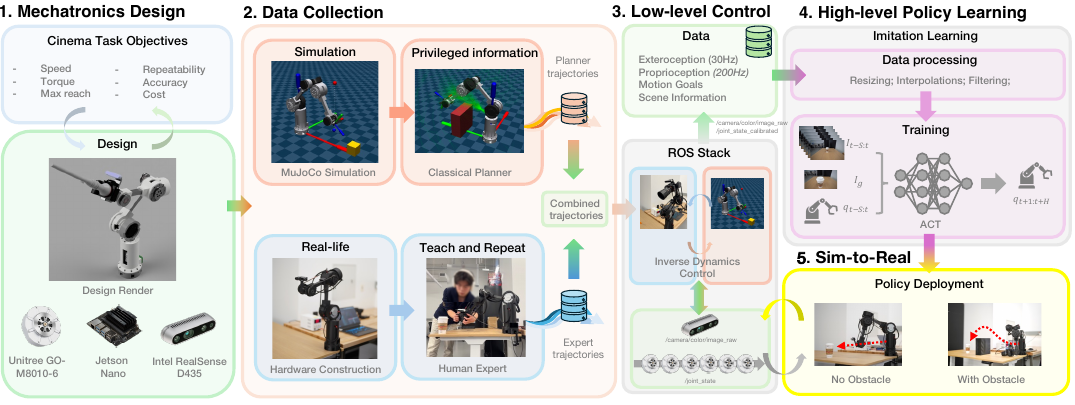}
    \caption{Overview of the IRIS system pipeline. Cinema task objectives guide task-specific hardware design. Training data are collected exclusively from real-world human demonstrations, while classical planner trajectories generated in simulation are used for analysis and comparison. A ROS-based low-level control stack executes inverse-dynamics control, and a goal-conditioned imitation learning policy (ACT) is trained on human data and deployed on the physical robot, enabling smooth, obstacle-aware cinematic motion via sim-to-real transfer.}
    \label{fig:pipeline}
    \vspace{-1em}
\end{figure*}

The remainder of the paper is organized as follows: we give an overview of the related work in 
Section~\ref{sec:related}.
Section~\ref{sec:system} presents the mechatronics design of IRIS.
Section~\ref{sec:sim} describes the integration of the system, including simulation, ROS stack, and low-level control.
The visuomotor imitation learning is presented in Section~\ref{sec:il}. 
Section~\ref{sec:experiment} presents experimental evaluations on the physical system.
Finally, Section~\ref{sec:conclusion} concludes the paper and outlines future work.
An overview of the complete system pipeline is shown in \autoref{fig:pipeline}.

\section{Related Work}
\label{sec:related}

In this section, we briefly discuss the cinema robot hardware and motion control methods relevant to our design goals.

%

\subsection{State-of-the-art Cinema Robots}
\label{sec:related-hardware}

Robotic camera systems are widely used in professional film production due to their ability to execute precisely choreographed, highly repeatable motion required for visual-effects continuity and multi-pass composition \cite{fielding1985technique, mrmc_faq_motion_control}.
Commercial state-of-the-art cinema robots, e.g. the Bolt High-Speed Cinebot and Kira Cinema Robot \cite{mrmc_faq_motion_control, mpstudio_kira}, achieve high speed, high payload capacity, and sub-millimeter repeatability, but rely on industrial-grade hardware that leads to high capital cost (often exceeding \$50,000\,USD) and operational complexity. 
These systems are also typically operated by trained specialists and remain inaccessible to independent creators, small studios, and research environments.

Accessible and open-source 3D-printed manipulators such as Arctos, and Moveo \cite{arctos, moveo} demonstrate the feasibility of low-cost robotic arms using 3D-printing.
But they often lack the workspace, speed, or sufficient payload required for cinematic camera motion.
Those stepper-motor-based designs with high gear-ratio introduce cogging, backlash, and limited control bandwidth. 
%
To address these limitations, recent manipulators explore quasi-direct-drive (QDD) and remote-actuation architectures that reduce distal inertia and enable back-drivable, low-impedance dynamics.
Systems such as OpenArm and Berkeley Blue \cite{openarm, gealy2019blue} employ low-ratio brushless motors to achieve smooth, expressive motion suitable for dynamic tasks.
\iris{} adopts these design principles through a QDD, belt-driven architecture tailored for camera payloads, speed, achieving millimeter-level repeatability while maintaining the smoothness required for cinematic camera tracking.
A detailed comparison of technical specifications between \iris and existing state-of-the-art cinematic manipulators is provided in \autoref{tab:cinema_arm_comparison}.

\subsection{Motion Control Methods}
\label{sec:related-software}
Most commercial cinema robots rely on classical motion generation techniques, including keyframe interpolation, time-parameterized splines, and inverse-kinematics-based trajectory replay \cite{mrmc_faq_motion_control}.
To handle obstacle avoidance and workspace constraints, these systems often incorporate sampling-, graph-, or optimization-based planners such as A*, RRT*, CHOMP, and TrajOpt \cite{hart1968astar, karaman2011rrtstar, ratliff2009chomp, schulman2013trajopt}.
While those methods produce deterministic and geometrically feasible trajectories, they optimize hand-crafted objectives and fail to capture perceptual qualities central to cinematography, such as pacing, framing stability, and expressive smoothness.

Learning-based methods offer an alternative by learning planning objectives directly from data.
%
Imitation learning (IL) enables end-to-end mapping from observations to actuator commands using expert demonstrations, and has been widely applied through behavior cloning (BC), dataset aggregation (DAgger), and generative adversarial formulations \cite{levine2016visuomotor, ross2011dagger, ho2016gail}.
However, vanilla IL methods are sensitive to distribution shift and compounding errors.
%
Recent transformer- and diffusion-based architectures address long-horizon temporal dependencies and compounding error by modeling temporally extended action sequences, enabling smooth, high-frequency control  \cite{zhao2023aloha, chi2023diffusionpolicy, janner2022planning}. 
%
Inverse reinforcement learning (IRL) and guided cost learning extend this paradigm by inferring underlying cost functions from demonstrations rather than manually specifying them \cite{ziebart2008maximum, finn2016guided}.
Reinforcement learning (RL) further generalizes through end-to-end policy optimization, but often requires large-scale interaction data and faces significant sim-to-real challenges in perception-driven control \cite{haarnoja2018soft, kober2013reinforcement}.
%

%
Building on these insights, \iris{} uses a goal-conditioned ACT formulation integrated with task-specific hardware and trained with camera expert demonstrations.
By framing camera motion as a visuomotor problem, the learned policy generates obstacle-aware, goal-directed trajectories directly from RGB observations \cite{chaumette2006visual}, eliminating the need for pre-mapped environments or hand-crafted motion costs.

\section{Mechatronics Design}
\label{sec:system}
\begin{figure}[t]
    \centering
    \includegraphics[width=0.7\linewidth]{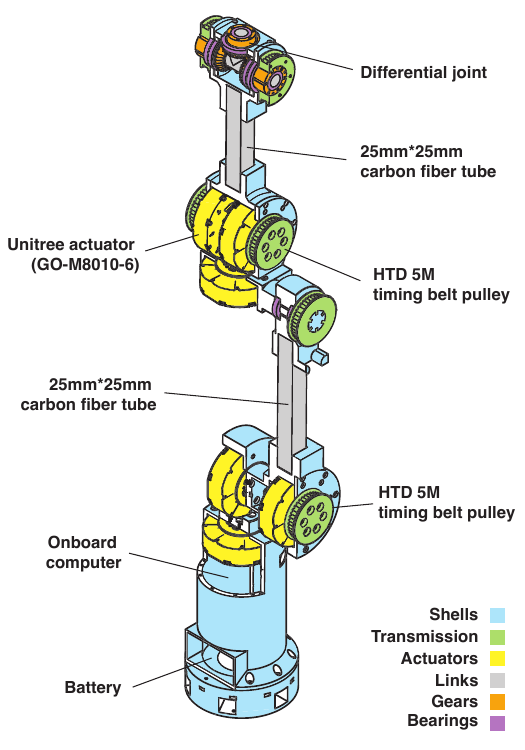}
    \caption{\iris{} hardware overview: lightweight task-specific architecture with relocated actuation and a differential wrist.}
    \label{fig:hardware}
    \vspace{-1em}
\end{figure}

\begin{table*}[ht]
\centering
\caption{Comparison of commercial cinema robots and 3D-printed robot arms for cinematic applications.}
\vspace{-3mm}
\label{tab:cinema_arm_comparison}

\footnotesize
\setlength{\tabcolsep}{4pt}

\begin{tabular}{lccccccc}
\toprule
\textbf{Robot Arm} &
\textbf{Reach} &
\textbf{Speed} &
\textbf{Acceleration} &
\textbf{Repeatability} &
\textbf{Payload} &
\textbf{Ease of Use} &
\textbf{Cost} \\
 & (mm) & (m/s) & (m/s$^2$) & (mm) & (kg) &  & (USD) \\
\midrule

\textbf{Commercial Cinema Robots} \\
\midrule
\rowcolor{bestcommercial}
Bolt Cinebot~\cite{mrmc_faq_motion_control} &
\cellcolor{appleorange}2000 &
5.0 &
30 &
$\pm$0.05 &
\cellcolor{appleorange}20 &
\cellcolor{appleorange}Medium &
\$69k+ \\

Colossus~\cite{motorizedprecision_colossus} &
1100 &
\cellcolor{appleorange}6.8 &
\cellcolor{appleorange}40 &
$\pm$0.05 &
10+ &
Medium &
\$200k+ \\

Kira~\cite{mpstudio_kira} &
1500 &
3.5 &
20 &
$\pm$0.05 &
10 &
Low &
\$50k+ \\

E-Jib Mini~\cite{lbb_moco_guide} &
1800 &
1.0 &
5 &
$\pm$0.1 &
5 &
Low &
\cellcolor{appleorange}\$9k \\

\midrule
\textbf{3D-Printed / Open-Source Arms} \\
\midrule
\rowcolor{best3d}
Dexter~\cite{dexter} &
700 &
0.5 &
2 &
\cellcolor{appleorange}$\pm$0.025 &
\cellcolor{appleorange}3 &
Low &
$\sim$\$2k \\

Arctos~\cite{arctos} &
600 &
0.4 &
1.5 &
$\pm$5 &
0.5 &
Low &
$\sim$\$1.5k \\

BCN3D Moveo~\cite{moveo} &
625 &
0.3 &
1 &
$\pm$0.1 &
0.5 &
Low &
$\sim$\$1k \\

Mirobot~\cite{wlkata_website} &
400 &
0.2 &
0.5 &
$\pm$0.2 &
0.5 &
High &
$\sim$\$1.6k \\

Blue~\cite{gealy2019blue} &
840 &
2.1 &
-- &
$\pm$3.7 &
2 &
Medium &
$\sim$\$5k \\

\midrule
\rowcolor{bestoverall}
\textbf{This Work (\iris{})} &
\cellcolor{appleorange}940 &
\cellcolor{appleorange}3.3 &
\cellcolor{appleorange}15 &
$\pm$1 &
1.5 &
\cellcolor{appleorange}High &
\cellcolor{appleorange}$\sim$\$0.95k \\

\bottomrule
\end{tabular}

\vspace{0.4em}
{\footnotesize
\colorsquare{appleorange} Best value per metric across all systems
(maximum for reach, speed, acceleration, and payload; minimum for repeatability and cost).
\;
\colorsquare{best3d} Best-performing 3D-printed open-source robot arm.
\;
\colorsquare{bestcommercial} Best-performing commercial cinema robot.
\;
\colorsquare{bestoverall} Best overall balance of cinematic performance, usability, and cost.
}
\vspace{-1em}
\end{table*}
\iris{} is a vertically integrated cinema robot comprising a task-specific 6-DOF manipulator with brushless-DC (BLDC) actuators, joint-space impedance control, and an end-effector-mounted Intel RealSense camera D435 \cite{intel_realsense_d435}.
As shown in \autoref{fig:hardware}, the hardware is co-designed with the task objectives to prioritize long reachability, high repeatability, and high accuracy while remaining low-cost, compact, and portable.
%

\subsection{Mechanical Structure}
\label{sec:mech}
The mechanical design of \iris{} prioritizes low distal inertia to improve dynamic responsiveness and achieve smooth precise camera motion.
The arm provides approximately 1\,m of reach, supports up to 1.5\,kg payload, and has a total system mass of approximately 8.5\,kg.
The complete hardware system costs under \$1{,}000\,USD in materials, including the embedded computer and the camera unit.
%
%
To reduce reflected inertia at the end effector, actuators are designed to be concentrated near the base.
The elbow pitch joint employs a HTD-5M timing belt transmission, enabling proximal motor placement.
The wrist design is achieved via a two-motor differential, driving pitch and roll actuation.
This design eliminates wrist-mounted motors, reducing total distal inertia at the wrist.
The mechanical structure explicitly decouples translational and rotational DOF, reducing kinematic coupling and allowing faster and more stable inverse kinematics (IK) solutions.
The robot links consist of two 20$\times$20\,mm carbon-fiber tube links and 24 custom 3D-printed components, including joint housings, timing-belt pulleys, and structural supports.
All non-standard structural components are fabricated using 3D-printing, enabling accessible, low-cost reproduction without specialized machining.

\subsection{Electronics}
The system contains six actuators daisy-chained over a half-duplex RS-485 bus, interfaced to the host computer via a USB-to-serial adapter.
While \iris{} supports embedded operation on an NVIDIA Jetson Nano, experiments were conducted on a laptop with an NVIDIA RTX 4080 GPU for low-latency inference.
Power is supplied by a 3S Li-Po battery (24\,V nominal) or regulated DC input, with an average draw of 4 A (around 100\,W) and 1\,A at idle. 
It peaks up to 10\,A during high-dynamic motion.

\subsection{Actuation and Control Stack}
\label{sec:control}
To address the conflicting requirements of high-speed and high accuracy, we use Quasi-Direct Drive (QDD) for our actuation \cite{gealy2019quasidirect}. 
QDD offers high torque density and backdrivability while maintaining low costs (approx. \$80\,USD per unit for our educational priced actuator).
\iris{} is actuated by Unitree GO-M8010-6 BLDC motors \cite{unitree_m8010_manual}, featuring a 6.33:1 planetary reduction and integrated field-oriented control (FOC).
The actuators operate at a nominal 24\,V (12-30\,V range), delivering peak torques of 23.7\,N$\cdot$m and velocities up to 30\,rad/s.
Each 530\,g unit provides 15-bit absolute position sensing, with real-time feedback for velocity and estimated torque.
Our low-level control architecture decouples high-level planning from a 200\,Hz impedance control loop~\cite{wensing2018proprioceptive}.
The pipeline streams joint positions, velocities, torque targets, and impedance gains through the vendor software development kit (SDK), enabling compliant yet precise motion execution.
To ensure trajectory smoothness, commands are pre-processed through a first-order low-pass filter ($\alpha=0.08$) and velocity-limited ramping (0.6-1.0\,rad/s). 
For safety, a 250\,ms command timeout triggers an automatic transition to a passive state estimation mode.
\section{System Integration}
\label{sec:sim}

\subsection{MuJoCo Simulation}
\begin{figure}[ht]
    \centering
    \includegraphics[width=\linewidth]{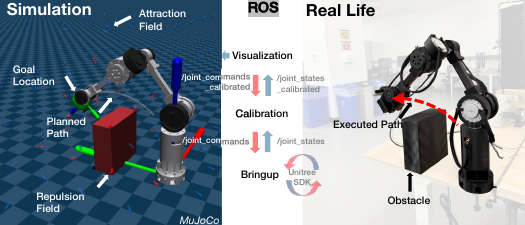}
    \caption{Sim-to-real execution of planner-generated trajectories. A classical potential-field planner generates collision-free reference paths in simulation (left), which are then executed on the physical IRIS robot (right) via a ROS-based control stack for validation and comparison.}
    \vspace{-1em}
    \label{fig:simulation}
\end{figure}

A high-fidelity MuJoCo simulation of \iris{} is developed for safe policy evaluation and trajectory visualization (\autoref{fig:simulation}).
The MuJoCo model captures \iris{}'s accurate kinematics, inertial properties, and collision geometry to support motion planning validation.
It also enables systematic evaluation of classical path-planning baselines, including sampling-based rapidly-exploring random tree star (RRT*), prior to real-world deployment.
%
To minimize the sim-to-real gap, we adopt similar actuator modeling protocols similar to the Unitree GO2 \cite{unitree_go2}, Disney Research's original bipedal robot \cite{disney_biped} and their recent Olaf \cite{müller2025olafbringinganimatedcharacter}, which all use the same actuator as \iris. 
Joint parameters (damping, armature inertia, friction) are tuned to match empirical step responses from the physical system, resulting in reduced sim-to-real gap.

\subsection{ROS Interface and Calibration}
\label{sec:ros_calib}
We develop a custom ROS package for \iris{} that bridges the Unitree SDK, MuJoCo, and the learning stack, as seen in \autoref{fig:simulation}.
A hardware bringup node manages low-level actuator I/O through the \texttt{/joint\_states} and \texttt{/joint\_commands} topics. 
Since the actuators use incremental encoders, a startup homing sequence is required to initialize \iris{} by placing it in a vertical upright pose to establish absolute joint references. 
Another calibration node maps raw encoder readings to kinematic coordinates (\texttt{\_calibrated}) using the homing offsets and does the differential wrist transformation.
The 200\,Hz bringup node synchronizes real-time state feedback with the MuJoCo model and applies velocity-limited commands, enabling reproducible sim-to-real deployment.


\subsection{Inverse Kinematics}
\label{sec:ik}
We use a numerical Jacobian-based IK solver to handle redundancy and singularities without requiring an analytical formulation.
At each control step, the Cartesian pose error is
$e = [\, p_t - p,\; r_t - r \,]$,
where $p,r \in \mathbb{R}^3$ are the current end-effector position and orientation, respectively, and $p_t,r_t$ are the corresponding targets.
Joint updates are computed via damped least squares,
$\Delta q = J^\top (J J^\top + \lambda I)^{-1} e$,
where $J \in \mathbb{R}^{6 \times n}$ is the Jacobian, $n$ the number of joints, and $\lambda>0$ the damping coefficient.
An intermediate joint estimate is computed as
$q_{\mathrm{ik}} = q + \eta \Delta q$,
where $q \in \mathbb{R}^n$ is the current joint state and $\eta$ is a step size.
The command is smoothed using an exponential moving average, $q_{\mathrm{cmd},t} = \alpha\, q_{\mathrm{ik},t} + (1-\alpha)\, q_{\mathrm{cmd},t-1}$, where $\alpha \in (0,1)$ is the smoothing coefficient and $q_{\mathrm{cmd},t}$ denotes the final commanded joint configuration at time $t$.
The filtered command $q_{\mathrm{cmd}}$ is applied to the controller and used to initialize the next IK iteration.

\section{Imitation Learning for Motion Control}
\label{sec:il}
\begin{figure*}
    \centering
    \includegraphics[width=\linewidth]{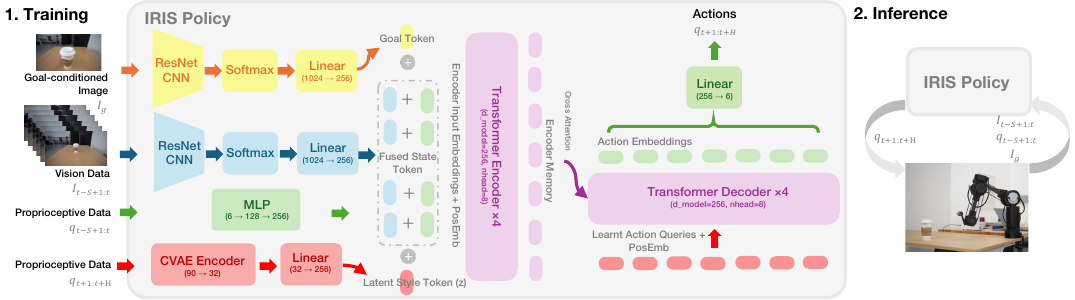}
    \caption{IRIS policy architecture. During training (left), the model conditions on observation history, a goal image, and a CVAE-encoded latent style token $z$ derived from the ground-truth future trajectory. Visual inputs pass through a shared ResNet-18 and Spatial Softmax to preserve spatial coordinates, then fuse with proprioception into temporal tokens. A transformer decoder predicts the 15-step joint trajectory $\hat{q}_{t+1:t+H}$. At inference (right), the CVAE branch is replaced with $z=0$ for deterministic execution.}
    \label{fig:architecture}
    \vspace{-1em}
\end{figure*}

To enable intuitive and autonomous cinematic control, we introduce a learning-based framework where users specify shots via a single target image rather than explicit geometric waypoints.
Our design uses a goal-conditioned adaptation of ACT \cite{zhao2023aloha}, augmented with a conditional variational autoencoder (CVAE). 
This architecture explicitly models the multimodal distribution of expert cinematic styles, allowing for diverse valid trajectories. 
By formulating camera motion as an IL problem, the learning pipeline is able to generate human-like, obstacle-aware trajectories that converge to the desired framing.
\autoref{fig:architecture} provides an overview of the policy architecture.

\subsection{Problem Formulation}
The camera motion control problem is formulated as a goal-conditioned partially observable Markov decision process (POMDP).
At time step $t$, the observation is defined as
\begin{equation}
    o_t = (I_t, q_t),
\end{equation}
where $I_t \in \mathbb{R}^{H \times W \times 3}$ denotes the RGB image captured by the end-effector-mounted camera at time $t$, and
$q_t \in \mathbb{R}^{6}$ represents the robot joint configuration.
The task goal is specified by a \emph{goal image} $I_g \in \mathbb{R}^{H \times W \times 3}$ that encodes the desired visual framing.
To capture the multimodality of feasible cinematic motions, we introduce a latent variable $z \in \mathbb{R}^{d_z}$ and model the policy as a conditional variational distribution
\begin{equation}
    \hat{q}_{t+1:t+H} = \pi_\theta\!\left(o_{t-S+1:t},\, I_g,\, z\right),
\end{equation}
where $\pi_\theta$ is a parameterized policy with learnable parameters $\theta$, 
$o_{t-S+1:t} = \{o_{t-S+1}, \dots, o_t\}$ denotes a history of $S$ observations, and
$\hat{q}_{t+1:t+H} = \{\hat{q}_{t+1}, \dots, \hat{q}_{t+H}\}$ is the predicted sequence of absolute joint configurations over a horizon of $H$ time steps.
%

\vspace{-1em}
\subsection{Policy Architecture}
We parameterize $\pi_\theta$ using a DETR-style ACT~\cite{zhao2023aloha} augmented with a CVAE to capture diverse cinematic styles. 
A frozen ResNet-18 backbone encodes the observation history and goal image into spatial feature maps, which are processed via spatial softmax to explicitly preserve feature coordinates. 
These visual features are fused with proprioceptive embeddings at each timestep to form \emph{fused state tokens}. 
During training, a CVAE encoder maps the ground-truth trajectory to a latent variable $z$, which is projected into a \emph{latent style token} and prepended to the transformer input as \texttt{[Style, State, Goal]}. 
At deployment, $z$ is replaced with a zero vector for deterministic execution.
The transformer encoder processes this multimodal context, enabling the decoder to predict future joint trajectories $\hat{q}_{t+1:t+H}$ via cross-attention. 
The complete architecture comprises 19.1M parameters, with 11.2M in the frozen backbone and a 7.9M trainable policy, of which 93\% (7.4M) are in the transformer layers.
The remaining parameters are distributed between the input layers (404k) and the training-only CVAE encoder (105k).

\begin{algorithm}[t]
\caption{IRIS ACT--CVAE Policy Training}
\label{alg:iris_act_train}
\small
\begin{algorithmic}[1]
\Require Dataset $\mathcal{D}$, history length $S$, horizon $H$
\State Initialize policy parameters $\theta$ and CVAE encoder $\phi$
\State Initialize optimizer (AdamW)
\For{each training iteration}
    \State Sample $(I_{t-S+1:t},\, q_{t-S+1:t},\, I_g,\, q^*_{t+1:t+H}) \sim \mathcal{D}$
    \State Sample latent $z \sim q_\phi(z \mid q^*_{t+1:t+H})$
    \State $\hat{q}_{t+1:t+H} \gets 
    \pi_\theta(I_{t-S+1:t},\, q_{t-S+1:t},\, I_g,\, z)$
    \State Update $(\theta,\phi)$ using loss 
    $\mathcal{L}$ from Eq.~\eqref{eq:total_loss}
\EndFor
\end{algorithmic}
\end{algorithm}


\begin{algorithm}[t]
\caption{IRIS ACT--CVAE Policy Deployment}
\label{alg:iris_deploy}
\small
\begin{algorithmic}[1]
\Require Trained policy $\pi_\theta$, goal image $I_g$
\State Initialize observation buffer of length $S$
\State Initialize ROS I/O
\While{ROS is running}
    \State Receive $(I_t, q_t)$ and append to buffer
    \If{buffer not full}
        \State \textbf{continue}
    \EndIf
    \State $\hat{q}_{t+1:t+H} \gets 
    \pi_\theta(I_{t-S+1:t},\, q_{t-S+1:t},\, I_g,\, z{=}0)$
    \State $q^{\mathrm{cmd}} \gets \mathrm{EMA}(\hat{q}_{t+1})$
    \State Publish $q^{\mathrm{cmd}}$ at control rate $f_c$
\EndWhile
\end{algorithmic}
\end{algorithm}

\subsection{Loss Function}
The policy is trained by minimizing a weighted composite objective $\mathcal{L}$, which is given by
\begin{equation}
\label{eq:total_loss}
\mathcal{L}
=
\mathcal{L}_{\mathrm{mse}}
+
\mathcal{L}_{\mathrm{kl}}
+
\mathcal{L}_{\mathrm{smooth}}.
\end{equation}
\vspace{-1em}

In this objective, $\mathcal{L}_{\mathrm{mse}}$ is defined as
\begin{equation}
    \mathcal{L}_{\mathrm{mse}}=\frac{1}{H}\sum_{i=1}^{H}\|\hat{q}_{t+i}-q^*_{t+i}\|_2^2,
\end{equation}
enforcing accurate trajectory tracking in joint space via reconstruction loss,
$\mathcal{L}_{\mathrm{kl}}=\beta\,D_{\mathrm{KL}}\!\left(q_\phi(z\mid\cdot)\,\|\,\mathcal{N}(0,I)\right)$ regularizes the variational latent distribution using KL divergence,
and
\begin{equation}
    \mathcal{L}_{\mathrm{smooth}}=\lambda\sum_{i=2}^{H}\|\hat{q}_{t+i}-\hat{q}_{t+i-1}\|_1,
\end{equation}
penalizes high-frequency temporal variations in the predicted joint configurations to ensure smoothness.



\subsection{Data Collection}
\label{sec:dataset}
\begin{figure}[t]
    \centering
    \includegraphics[width=\linewidth]{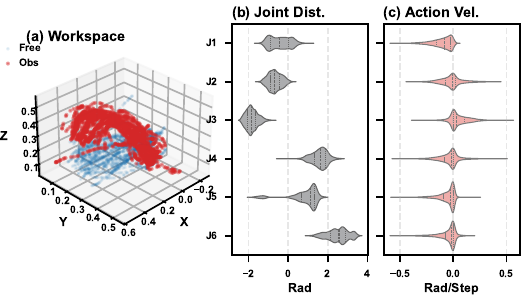}
    \caption{Demonstration dataset coverage.
(a) End-effector workspace coverage over all demonstrations.
(b) Joint-space distributions of the six joints.
(c) Per-step joint displacement magnitude, indicating bounded control actions.}
    \label{fig:data}
    \vspace{-1em}
\end{figure}

Since cinematic nuances are difficult to model mathematically, we use camera experts to teach directly on the physical \iris hardware. 
An operator manually guides the end-effector to produce a push-in shot tracking a coffee cup in the zero-torque actuation mode.
We collect synchronized demonstration data via ROS, recording joint states at 200\,Hz and RGB images at 30\,Hz.
Demonstrations are segmented into complete shot episodes and converted into training clips using a sliding window with observation horizon $S{=}8$ and prediction horizon $H{=}15$, where the final frame of each episode serves as the goal image.
The dataset consists of 132 episodes and 13,954 clips, balanced between unobstructed (6,680 clips) and obstacle-avoidance (7,274 clips) push-in shots.
Episodes are split at the episode level into train/validation/test sets (80/10/10), yielding 11,039 training, 1,306 validation, and 1,609 test clips.
\autoref{fig:data} visualizes the dataset distribution and expert trajectories.


\subsection{Training}
\label{sec:training}

All imitation learning policies are trained 
%
with batch size 64 on a single NVIDIA RTX~4090 GPU (Alienware X16, Intel i9, Ubuntu~22). 
The complete hyperparameters used during training can be seen in \autoref{tab:hyperparams}.
Models are trained on 11,039 training clips and validated on 1,306 validation clips. 
All RGB inputs are resized to $224\times224$ and normalized using ImageNet statistics before the training loop to reduce the CPU bottleneck speed down.
This pre-processing step reduces per-epoch training time from over 20 minutes to under 5 minutes.
Each epoch consists of 345 gradient update steps, resulting in approximately 34,500 total updates over 100 epochs. 
On the RTX4090, one epoch takes about 300 seconds, resulting in a total training time of roughly 8 hours.

\begin{table}[t]
\centering
\caption{Key model and training hyperparameters.}
\vspace{-3mm}
\label{tab:hyperparams}
\setlength{\tabcolsep}{6pt}
\renewcommand{\arraystretch}{1.15}
\begin{tabular}{l c}
\toprule
\textbf{Parameter} & \textbf{Value} \\
\midrule
Observation horizon $S$          & 8 \\
Prediction horizon $H$           & 15 \\
Visual backbone                  & ResNet-18 (ImageNet-pretrained) \\
Image resolution                 & $224 \times 224$ \\
Transformer dimension $d$        & 256 \\
Attention heads                  & 8 \\
Encoder / decoder layers         & 4 / 4 \\
Latent dimension $z$             & 32 \\
\midrule
Batch size                       & 64 \\
Optimizer                        & AdamW \\
Learning rate                    & $1\times10^{-4}$ \\
Training epochs                  & 100 \\
KL weight $\beta$                & 0.01 \\
Smoothness weight $\lambda_{\text{smooth}}$ & 0.01 \\
\bottomrule
\end{tabular}
\vspace{-1em}
\end{table}

\subsection{Policy Deployment}
\label{sec:deployment}
We deploy the policy on an ROS-based controller running at 10Hz, using a synchronizer to align live RGB images and joint states into a history buffer of length $S=8$. 
At each control step, the model predicts a trajectory of future absolute joint positions $\hat{q}_{t:t+H}$. 
We employ a receding-horizon strategy with a lookahead of $k=1$ to compensate for system latency.
Before actuating, the raw target $\hat{q}_{t+k}$ is first clamped to a maximum deviation of $\delta_{\text{max}}=0.2$\,rad from the current physical state. 
Subsequently, an exponential moving average (EMA) filter with $\alpha=0.3$ is applied to suppress high-frequency jitter, calculated as $q^{\text{cmd}}_{t+1} = \alpha \hat{q}_{t+k} + (1-\alpha) q^{\text{cmd}}_{t}$. 
Inference is performed on a single NVIDIA RTX 3070 with a forward-pass latency less than 10\,ms.

\section{Experiments}
\label{sec:experiment}
The evaluation comprises two components: a low-level control study to assess the intrinsic motion accuracy of \iris{}, 
and a high-level learning-based motion control evaluation to validate the effectiveness of goal-conditioned shot execution. 
%
All experiments are conducted on the physical \iris{} robot under identical workspace layouts and configurations.





\subsection{Low-Level Control Evaluation}
\label{sec:low_level}

\begin{figure}[t!]
    \centering
    \includegraphics[width=\linewidth]{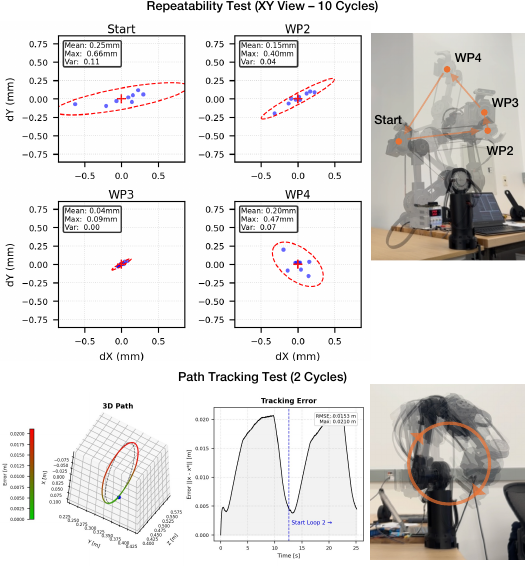}
    \caption{
    Low-level evaluation. Top: End-effector XY scatter over $K=10$ trials at start and WP2–WP4. Bottom: Reference vs. executed 3D trajectories and Cartesian error. \iris{} achieves sub-millimeter repeatability and centimeter-scale accuracy.
    }
    \vspace{-1em}
    \label{fig:repeat}
\end{figure}

We first validate the robot's hardware fidelity to ensure it supports learning-based control; see \autoref{fig:repeat}.

\textit{Repeatability.}
We repeatedly execute a fixed joint-space trajectory for $K=10$ trials, stopping at four distinct 3D waypoints, and evaluate repeatability at each point as the mean deviation from the centroid. 
We define the repeatability as 
\begin{equation}
    E_{\mathrm{rep}} = \frac{1}{K}\sum \|\mathbf{x}_T^k - \bar{\mathbf{x}}_T\|_2 .
\end{equation}

\textit{Tracking Accuracy.}
Path-tracking accuracy is evaluated by executing a predefined circular trajectory in 3D space for two cycles and comparing the desired end-effector trajectory with the executed trajectory reconstructed from joint encoder measurements.
We formulate the tracking error using the root mean square error (RMSE)
\begin{equation}
    E_{\mathrm{track}} = \frac{1}{T} \int \left\| \mathbf{x}(t) - \mathbf{x}^*(t) \right\|_2^2 \, dt .
\end{equation}
%

\subsection{Shot Automation Experiment Setup}
\label{sec:shot_tasks}
We evaluate the learning-based policy’s style reproduction and generalization on two tasks:
(i) \emph{Push-in (Unobstructed)}, a dolly-in toward a target (coffee cup) guided by a goal image; 
and (ii) \emph{Push-in (Obstacle Avoidance)}, which adds a cubic obstacle ($10\times25\times25$\,cm) to the path.
Each trial begins from a different initial pose and is repeated 10 times to assess shot quality and success rate. 
The experiment setup can be referred to \autoref{fig:qualitative} on the left.

\subsection{Benchmarks}
\label{sec:benchmarks}
We compare our approach against three state-of-the-art motion control strategies. Namely, 
(i) \textit{Human Expert Replay}, which utilizes a direct "Teach and Repeat" replay of expert demonstrations; 
(ii) \textit{Classical Planner (RRT*)}, a geometric baseline generating collision-free open-loop trajectories by implementing joint-space RRT*~\cite{karaman2011rrtstar} in MuJoCo with capsule-based collision checking (7.5\,cm safety radius) for offline planning between IK-solved configurations; 
and (iii) \textit{Vanilla BC}, a deterministic vanilla BC baseline~\cite{levine2016endtoend} that directly regresses future absolute joint positions from RGB observations, goal images, and joint history using a Multi-Layer Perceptron (MLP) and ResNet backbone for feature extraction.
\subsection{Evaluation Metrics}
\label{sec:metrics}
The system is evaluated on six core metrics that quantify cinematic quality and control precision. 
%
\begin{itemize}
    \item \textit{Visual Alignment ($\mathcal{S}_{\text{vis}}$).} 
 Quantifies semantic similarity between the last ($I_{\rm last}$) and goal ($I_g$) images using feature embeddings $\phi(\cdot)$ \cite{bromley1993signature} from the penultimate layer of a pre-trained ResNet-18~\cite{reis2024realtimeflyingobjectdetection, reis2024yolov8}.
    \begin{equation}
        \mathcal{S}_{\text{vis}} = \frac{\phi(I_{\rm last}) \cdot \phi(I_g)}{\|\phi(I_{\rm last})\|_2 \|\phi(I_g)\|_2}
    \end{equation}

    \item \textit{Success Rate.} The percentage of trials where the robot reaches the target configuration without collision and achieves a visual alignment score $\mathcal{S}_{\text{vis}} > 0.85$ (aligned).

    \item \textit{Cartesian Smoothness ($\mathcal{J}_{\text{cart}}$).} 
    Measures physical stability via the magnitude of the end-effector jerk \cite{hogan1984minimum}. 
    We compute the end-effector position $\mathbf{x}_t \in \mathbb{R}^3$ from joint states using a MuJoCo-based forward kinematics solver and approximate the third-order time derivative:
    \begin{equation}
        \mathcal{J}_{\text{cart}} = \| \dddot{\mathbf{x}}_t \|_2 \approx \left\| \frac{\Delta^3 \mathbf{x}_t}{(\Delta t)^3} \right\|_2
    \end{equation}

    \item \textit{Framing Error ($\mathcal{E}_{\text{frame}}$).} 
    The Euclidean pixel distance between last image center $\mathbf{c}$ and the target object's centroid $\mathbf{p}$, detected in real-time via YOLOv8 (Nano) \cite{reis2024realtimeflyingobjectdetection}:
    \begin{equation}
        \mathcal{E}_{\text{frame}} = \| \mathbf{p} - \mathbf{c} \|_2
    \end{equation}

    \item \textit{Subject Retention Rate (SRR).} The percentage of the trajectory where the target object remains within the central 50\% of the camera frame \cite{chaumette2006visual}, ensuring consistent subject tracking.

    \item \textit{Inference Latency.} The average wall-clock time required for the policy to process visual observations and output motor commands.

\end{itemize}

\subsection{Ablation Studies}
We validate our architecture design through two ablation studies, with 10 trials for each policy. 

\paragraph*{Action Space Formulation}
We compare our proposed \emph{absolute} control strategy ($q^{\text{cmd}}_{t+1} = \hat{q}_t$) against an \emph{incremental} action output ($q^{\text{cmd}}_{t+1} = q^{\text{curr}}_t + \Delta \hat{q}_t$) to verify if direct absolute position regression yields better global consistency for cinematic trajectories.

\paragraph*{Input Modality}
We analyze input modalities importance by comparing the full configuration ($\mathcal{O}=\{I_{t-S:t}, q_{t-S:t}, I_g\}$) against two variants: 
(i) \emph{Visual-Only} ($\mathcal{O}=\{I_{t-S:t}, I_g\}$), removing proprioception; 
and (ii) \emph{RGB-Only} ($\mathcal{O}=\{I_{t-S:t}\}$), removing both goal conditioning and proprioception to test implicit task inference.
The ablation studies results can be found in \autoref{tab:combined_results}.



\definecolor{appleorange}{HTML}{F5A623} 
\definecolor{bestoverall}{HTML}{FFF8E1} 
\definecolor{headergray}{HTML}{F2F2F2} 


\begin{table*}[!t]
\centering
\caption{Performance Comparison and Ablation Studies. Evaluation of task metrics across 10 randomized trials.}
\label{tab:combined_results}
\vspace{-3mm}

\footnotesize
\setlength{\tabcolsep}{3pt}
\renewcommand{\arraystretch}{1.2} 

\begin{tabularx}{\textwidth}{@{} l Y Y Y Y Y Y @{}}
\toprule
\textbf{Method} & 
\textbf{Success} $\uparrow$ & 
\textbf{Vis. Align.} $\uparrow$ & 
\textbf{Cart. Jerk} $\downarrow$ & 
\textbf{Framing} $\downarrow$ & 
\textbf{SRR} $\uparrow$ & 
\textbf{Latency} $\downarrow$ \\
 & (\%) & ($\mathcal{S}_{\text{vis}}$) & (m$/$s$^{3}$) & ($\mathcal{E}_{\text{frame}}$ px) & (\%) & (ms) \\

\midrule
\rowcolor{headergray} \multicolumn{7}{l}{\textbf{Baselines}} \\
\midrule
\rowcolor{bestcommercial} Human Expert Replay                 & \textbf{90.0} & \textbf{0.874} & 3.64 & 105.7 & \textbf{67.1} & -- \\
Classical Planner (RRT*)            & 10.0 & 0.636 & \cellcolor{appleorange}\textbf{0.22} & 219.6 & 10.5 & -- \\

\midrule
\rowcolor{headergray} \multicolumn{7}{l}{\textbf{Ablation Studies}} \\
\midrule
w/ Incremental Action Space         & 0.0 & 0.636 & 0.83 & \cellcolor{appleorange}\textbf{103.1} &  31.9 & 9.1 \\
w/ RGB Input Only                   & 0.0 & 0.584 & 1.65 & 113.0 & 7.2 & \cellcolor{appleorange} 9.0 \\
w/ RGB + Goal (No Proprio)          & 0.0 & 0.536 & 1.59 & 162.1 & 7.3 & 9.2 \\

\midrule
\rowcolor{bestoverall} 
\textbf{ACT--CVAE (Ours)} & 
\cellcolor{appleorange}\textbf{90.0} & 
\cellcolor{appleorange}\textbf{0.847} & 
\textbf{0.61} & 
183.6 & 
\cellcolor{appleorange} 33.1 & 
9.2 \\

\bottomrule
\end{tabularx}

\vspace{0.4em}
\parbox{\textwidth}{\footnotesize\raggedright
\colorbox{appleorange}{\phantom{xx}} Best value per metric. \;
\colorbox{bestoverall}{\phantom{xx}} Proposed method. \;
Human Expert and Classical Planner baselines are computed offline; therefore, latency is not reported. \\
$\uparrow$ indicates higher is better; $\downarrow$ indicates lower is better.
}
\vspace{-2em}
\end{table*}

\begin{figure*}[ht]
    \centering
    \includegraphics[width=\linewidth]{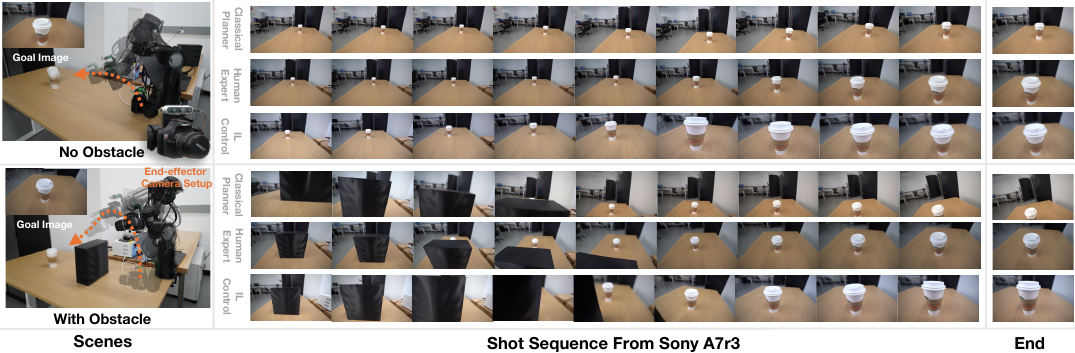}
    \caption{Qualitative shot reproduction results. Left: Experimental setup with the goal images for unobstructed (top) and obstacle (bottom) scenes. Right: Temporal image sequences comparing the Classical Planner, Human Expert, and our IL policy. Unlike the planner, which fails against unmodeled obstacles, our policy successfully maintains the desired framing while avoiding collisions, closely matching expert demonstrations.}
    \label{fig:qualitative}
    \vspace{-1em}
\end{figure*}





\subsection{Results and Discussion}
\label{sec:results}
The low-level control results validate the hardware integrity of \iris\ and the effectiveness of our low-level control architecture in maintaining the precision required for cinematic tasks. 
Evaluation of the system's mechanical consistency shows that \iris's repeatability remains sub-millimeter across all measured waypoints: 0.25\,mm (Start), 0.15\,mm (WP2), 0.04\,mm (WP3), and 0.20\,mm (WP4), with a maximum observed deviation of 0.66\,mm. 
Furthermore, the joint-space impedance controller achieves a RMSE of 1.53\,cm and a maximum error of 2.10\,cm. 
This performance indicates sufficient control fidelity and stability for reliable downstream visuomotor policy deployment in complex environments.



For the high-level policy experiments, the comparison metrics are summarized in \autoref{tab:combined_results}. 
Our ACT-CVAE policy achieves a 90\% success rate, significantly outperforming the classical planner (10\%). 
This performance gap highlights the advantage of closed-loop visuomotor control. 
While the open-loop planner fails against unmodeled sim-to-real gap, the learned policy dynamically adapts the trajectory and corrects the final pose to match the visual goal.
%
%
Additionally, our policy functions as a motion filter, producing trajectories significantly smoother than human experts ($0.61\,\text{m/s}^3$ vs. $3.64\,\text{m/s}^3$ jerk) by rejecting high-frequency jitter. 
This is achieved at the expense of responsiveness drops, with a lower SRR (33.1\%) compared to human operators (67.1\%). 
Despite this lag, it outperforms the classical planner (10.5\%), showing effective semantic object awareness where geometric baselines fail.
The qualitative shot results are further given in \autoref{fig:qualitative}.

Ablation studies show the importance of action space formulation, with incremental control failed completely (0\% success). 
We suspect that this is due to integration drift exacerbated by hardware compliance, whereas predicting \emph{absolute} joint positions provides the necessary self-correction for low-cost hardware. 
Furthermore, removing either proprioception or goal conditioning also resulted in total failure, verifying the necessity of full multimodal context.


Despite these capabilities, the broader robustness and functionalities of the system remain bounded by three factors: mechanical flexion during high-torque maneuvers, limited dataset diversity across scenes and objects, and a restricted repertoire of cinematic shot types. 
\section{Conclusions and Future Work}
\label{sec:conclusion}
We presented \iris, a low-cost, fully 3D-printed cinema robot system that integrates task-specific hardware with learning-based motion generation for automated shot execution. 
By coupling a cinematography-oriented mechanical design with visuomotor imitation learning, \iris achieves smooth and repeatable camera motion without reliance on expensive commercial platforms or expert-driven programming workflows. 
These results demonstrate that tightly designed and integrated hardware learning systems can make robotic cinematography more accessible and practical for a wider range of creators and research applications.
The current design can be extended in various aspects: at the hardware side, stiffening the chassis and optimizing actuator selection can minimize flexion. 
%
On the learning side, expanding the dataset to more diverse scenes and objects will enable real-time sequencing of multiple visual goals, supporting complex long-horizon camera motions such as compound dolly, crane, and pan trajectories.
Future work will explore these directions.
\bibliographystyle{IEEEtran} 
\bibliography{citations}  

\end{document}